\documentclass[10pt,twocolumn,letterpaper,conference]{article} 

\usepackage{avss}
\usepackage{times}
\usepackage{epsfig}
\usepackage{graphicx}
\usepackage{amsmath}
\usepackage{amssymb}
\usepackage{subcaption}
\usepackage{fancyhdr}
\usepackage{lipsum}

\avssfinalcopy 

\def\avssPaperID{80} 

\ifavssfinal\fi

\fancypagestyle{specialfooter}{%
	\fancyhf{}
	
	\fancyfoot[L]{978-1-5386-2939-0/17/\$31.00~\copyright~©2017 IEEE}
	\fancyhead[CO,CE]{To appear at International Conference on Advanced Video and Signal Based Surveillance (AVSS) 2017}
}

\begin{document}
	

	\title{CNN-based Cascaded Multi-task Learning of High-level Prior and Density Estimation for Crowd Counting}
	
	\author{Vishwanath A. Sindagi \quad Vishal M. Patel\\
		Department of Electrical and Computer Engineering, Rutgers University\\
		94 Brett Road, Piscataway, NJ, 08854, USA\\
		{\tt\small vishwanath.sindagi@rutgers.edu, vishal.m.patel@rutgers.edu}}
	

	\maketitle
	\thispagestyle{specialfooter}
	\begin{abstract}
		Estimating crowd count in densely crowded scenes is an extremely challenging task due to non-uniform scale variations. In this paper, we propose a novel end-to-end cascaded network of CNNs to jointly learn crowd count classification and density map estimation. Classifying crowd count into various groups is tantamount to coarsely estimating the total count in the image thereby incorporating a high-level prior into the density estimation network. This enables the layers in the network to learn globally relevant discriminative features which aid in estimating highly refined density maps with lower count error. The joint training is performed in an end-to-end fashion.  
		Extensive experiments on highly challenging publicly available datasets show that the proposed method achieves lower count error and better quality density maps as compared to the recent state-of-the-art methods. Furthermore, source code and pre-trained models are made available at https://github.com/svishwa/crowdcount-cascaded-mtl.
	\end{abstract}
	
	\section{Introduction}
	
	\setlength{\abovedisplayskip}{2pt}
	\setlength{\belowdisplayskip}{2pt}
	
	Crowd analysis has gained a lot of interest in recent years due to it's variety of applications such as video surveillance, public safety design and traffic monitoring.  Researchers have attempted to address various aspects of analyzing crowded scenes such as counting \cite{chan2008privacy,chan2012counting,skaug2016end,idrees2013multi}, density estimation \cite{lempitsky2010learning,zhang2016single,zhang2015cross,pham2015count,wang2016fast,boominathan2016crowdnet}, segmentation \cite{kang2014fully}, behavior analysis \cite{shao2014scene}, tracking  \cite{rodriguez2011density}, scene understanding \cite{shao2015deeply} and anomaly detection \cite{rabiee2016novel}. In this paper, we specifically focus on the joint task of estimating crowd count and density map from a single image.

	\begin{figure}[t]
		\centering
		\begin{minipage}{1\linewidth}
			\centering
			\includegraphics[width=1\linewidth]{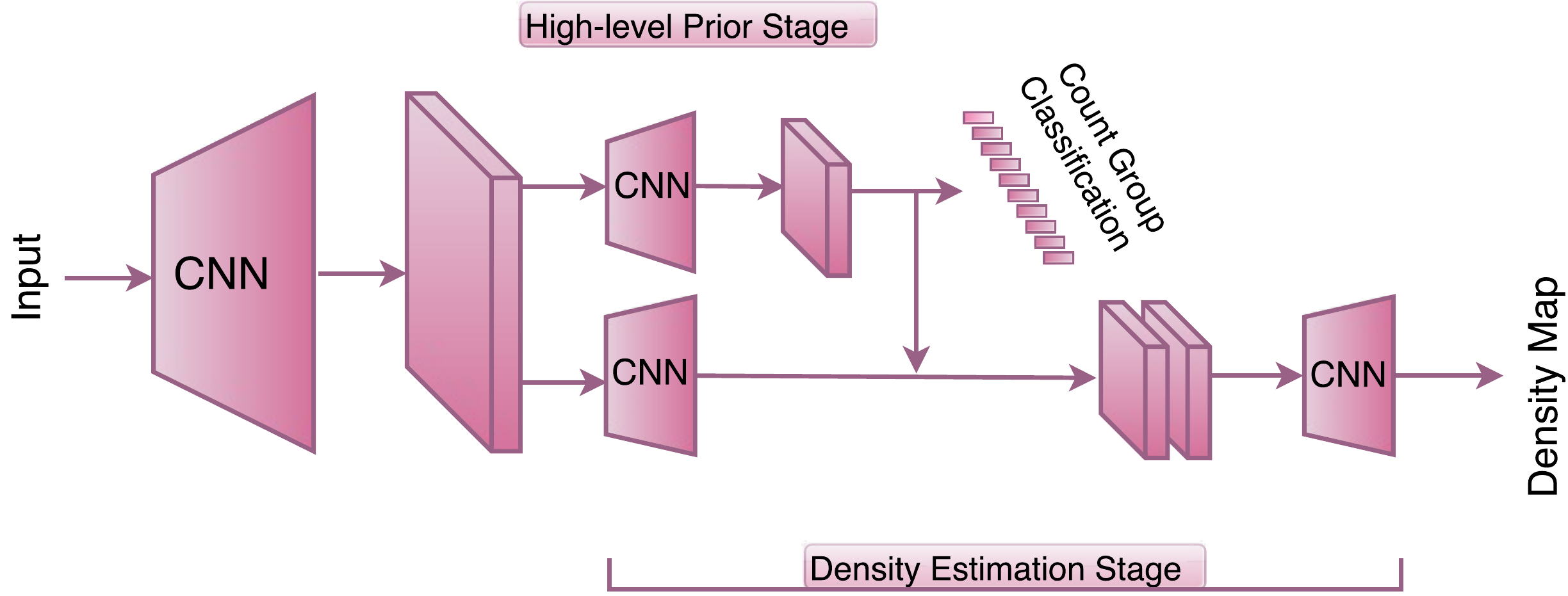}
			\captionsetup{labelformat=empty}
			\captionsetup{justification=centering}
			\caption*{(a)}
			\vspace{1pt}
			\vskip+6pt
		\end{minipage}
		\begin{minipage}{0.32\linewidth}
			\centering
			\includegraphics[width=1\linewidth]{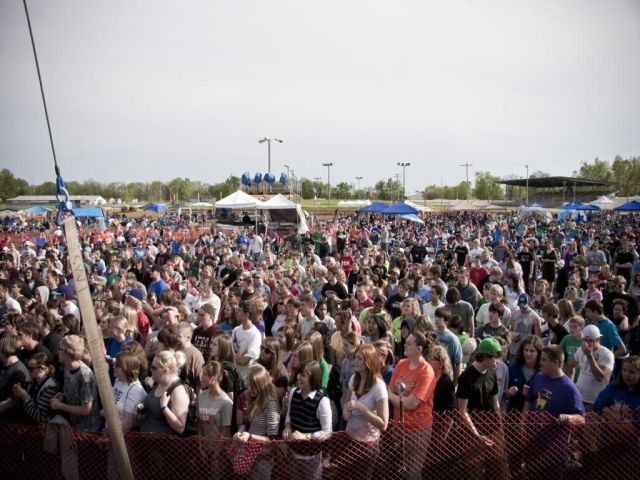}
			\captionsetup{labelformat=empty}
			\captionsetup{justification=centering}
			\caption*{(b)}
		\end{minipage}	
		\begin{minipage}{0.32\linewidth}
			\centering
			\includegraphics[width=1\linewidth]{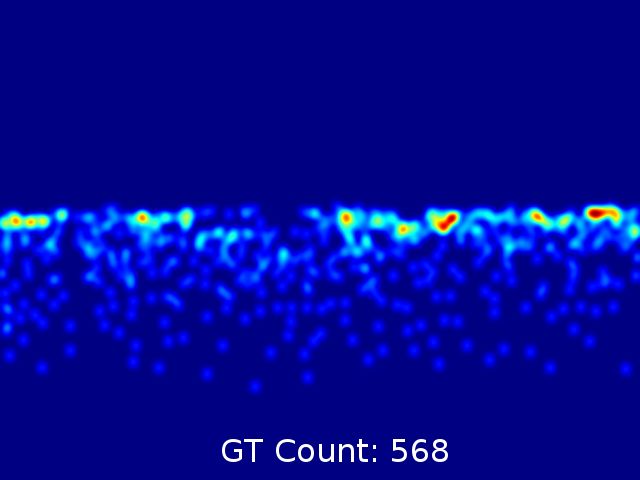}
			\captionsetup{labelformat=empty}
			\captionsetup{justification=centering}
			\caption*{(c)}
		\end{minipage}
		\begin{minipage}{0.32\linewidth}
			\centering
			\includegraphics[width=1\linewidth]{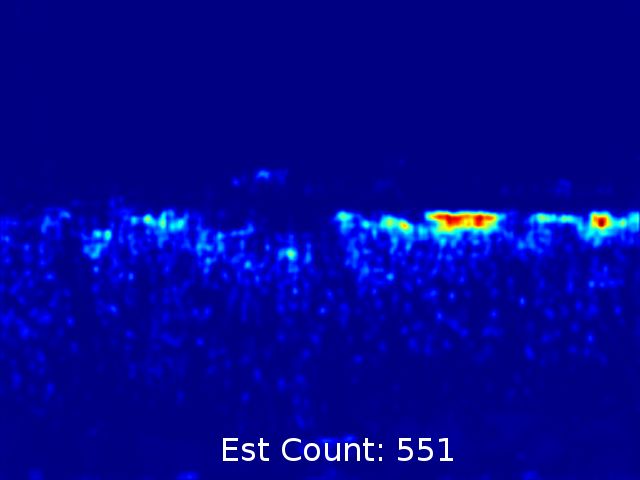}
			\captionsetup{labelformat=empty}
			\captionsetup{justification=centering}
			\caption*{(d)}
		\end{minipage}
		\vskip -8pt\caption{Proposed method and results. (a) Cascaded architecture for learning high-level prior and density estimation. (b) Input image (from the ShanghaiTech dataset \cite{zhang2016single}. (c) Ground truth density map. (d) Density map generated by the proposed method.} 
		\label{fig:firstcompare}
	\end{figure}

	One of the many challenges faced by researchers working on crowd counting is the issue of large variations in scale and appearance of the objects that occurs due to severe perspective distortion of the scene. Many methods have been developed that incorporate scale information into the learning process using different methods. Some of the early methods relied on multi-source and hand-crafted representations and catered only to low density crowded scenes \cite{idrees2013multi}. These methods are rendered ineffective in high density crowds and the results are far from optimal. Inspired by the success of Convolutional Neural Networks (CNNs) for various computer vision tasks, many CNN-based methods have been developed to address the problem of crowd counting
	\cite{boominathan2016crowdnet,bansal2015people,zhang2015cross}. Considering scale issue as a limiting factor to achieve better accuracies, certain CNN-based methods specifically cater to the issue of scale changes via multi-column  or multi-resolution network \cite{zhang2016single,onoro2016towards,skaug2016end}. 
	Though these methods demonstrated robustness to scale changes, they are still restricted to the scales that are used during training and hence are limited in their capacity to learn well-generalized models.

	The aim of this work is to learn models that cater to a wide variety of density levels present in the dataset by incorporating a high-level prior into the network. The high-level prior learns to classify the count into various groups whose class labels are based on the number of people present in the image. By exploiting count labels, the high-level prior is able to estimate coarse count of people in the entire image irrespective of scale variations thereby enabling the network to learn more discriminative global features. The high-level prior is jointly learned along with density map estimation using a cascade of CNN networks as shown in Fig. \ref{fig:firstcompare} (a). The two tasks (crowd count classification and density estimation) share an initial set of convolutional layers which is followed by two parallel set of networks that learn high-dimensional feature maps relevant to high-level prior and density estimation, respectively. The global features learned by the high-level prior are concatenated with the feature maps obtained from the second set of convolutional layers and further processed by a set of fractionally strided convolutional layers to produce high resolution density maps. Results of the proposed method on a sample input image are shown in Fig. \ref{fig:firstcompare} (c)-(d).

	\section{Related work}
	Traditional approaches for crowd counting from single images relied on hand-crafted representations to extract low level features. These features were then mapped to count or density map using various regression techniques. Loy \etal \cite{loy2013crowd} categorized existing methods into (1) detection-based methods (2) regression-based methods and (3) density estimation-based methods. 
	
	Detection-based methods typically employ sliding window-based detection algorithms to count the number of object instances in an image \cite{topkaya2014counting}. These methods are adversely affected by the presence of high density crowd and background clutter. To overcome these issues, researchers attempted to count by regression where they learn a mapping between features extracted from local image patches to their counts  \cite{ryan2009crowd,chen2012feature}. Using a similar approach, Idrees \etal \cite{idrees2013multi} fused count from multiple sources. The authors also introduced an annotated dataset (UCF\textunderscore CC\textunderscore 50) of 50 images containing 64000 humans. 
	
	Detection and regression methods ignore key spatial information present in the images as they regress on the global count. Hence, in order to incorporate spatial information present in the images, Lempitsky \etal \cite{lempitsky2010learning} introduced a new approach of learning a linear mapping between local patch features and corresponding object density maps. Instead of a linear mapping, Pham \etal in \cite{pham2015count} proposed to learn a non-linear function using a random forest framework. Wang and Zou \cite{wang2016fast} computed the relationship between image patches and their density maps in two distinct feature spaces. Recently, Xu and Qiu \cite{xu2016crowd} proposed to use much richer and extensive set of features for crowd density estimation. A more comprehensive survey of different crowd counting methods can be found in \cite{chen2012feature,li2015crowded}.
	

	More recently, due to the success of CNNs in various computer vision tasks, several CNN-based approaches have been developed for crowd counting \cite{wang2015deep,zhang2015cross,marsden2016fully,marsden2017resnetcrowd}.  Walach \etal \cite{walach2016learning} used CNNs with layered training approach.  In contrast to the existing patch-based estimation methods, Shang \etal \cite{skaug2016end} proposed an end-to-end estimation method using CNNs by simultaneously learning local and global count on the whole sized input images. Observing that the existing approaches cater to a single scale due to their fixed receptive fields, Zhang \etal \cite{zhang2016single} proposed a multi-column architecture to extract features at different scales. In addition, they also introduced a large scale annotated dataset (ShanghaiTech dataset). Onoro-Rubio  and L{\'o}pez-Sastre in \cite{onoro2016towards} addressed the scale issue by proposing a scale aware counting model called Hydra CNN. Boominathan \etal in \cite{boominathan2016crowdnet} proposed to tackle the issue of scale variation using a combination of shallow and deep networks along with an extensive data augmentation by sampling patches from multi-scale image representations.

	Zhang \etal \cite{zhang2016single} and Onoro \etal \cite{onoro2016towards} demonstrated that designing networks that are robust to scale variations is crucial for achieving better performance as compared to other CNN-based approaches. However, these methods rely on architectures that cater to selected set of scales thereby limiting their abilities to learn more generalized models.  Additionally, the recent approaches individually regress either on crowd count or density map. Among the approaches that estimate density maps, the presence of pooling layers in the existing approaches reduce the resolution of the output density map prohibiting one to regress on full resolution density maps. This results in the loss of crucial details especially in images containing large variation in scales. Considering these drawbacks, we present a novel end-to-end cascaded CNN network that jointly learns a high-level global prior and density estimation. The high-level prior enables the network to learn globally relevant and discriminative features that aid in estimating density maps from images with large variations in scale and appearance.  

	\begin{figure}[t!]
		\begin{center}
			\includegraphics[width=1\linewidth]{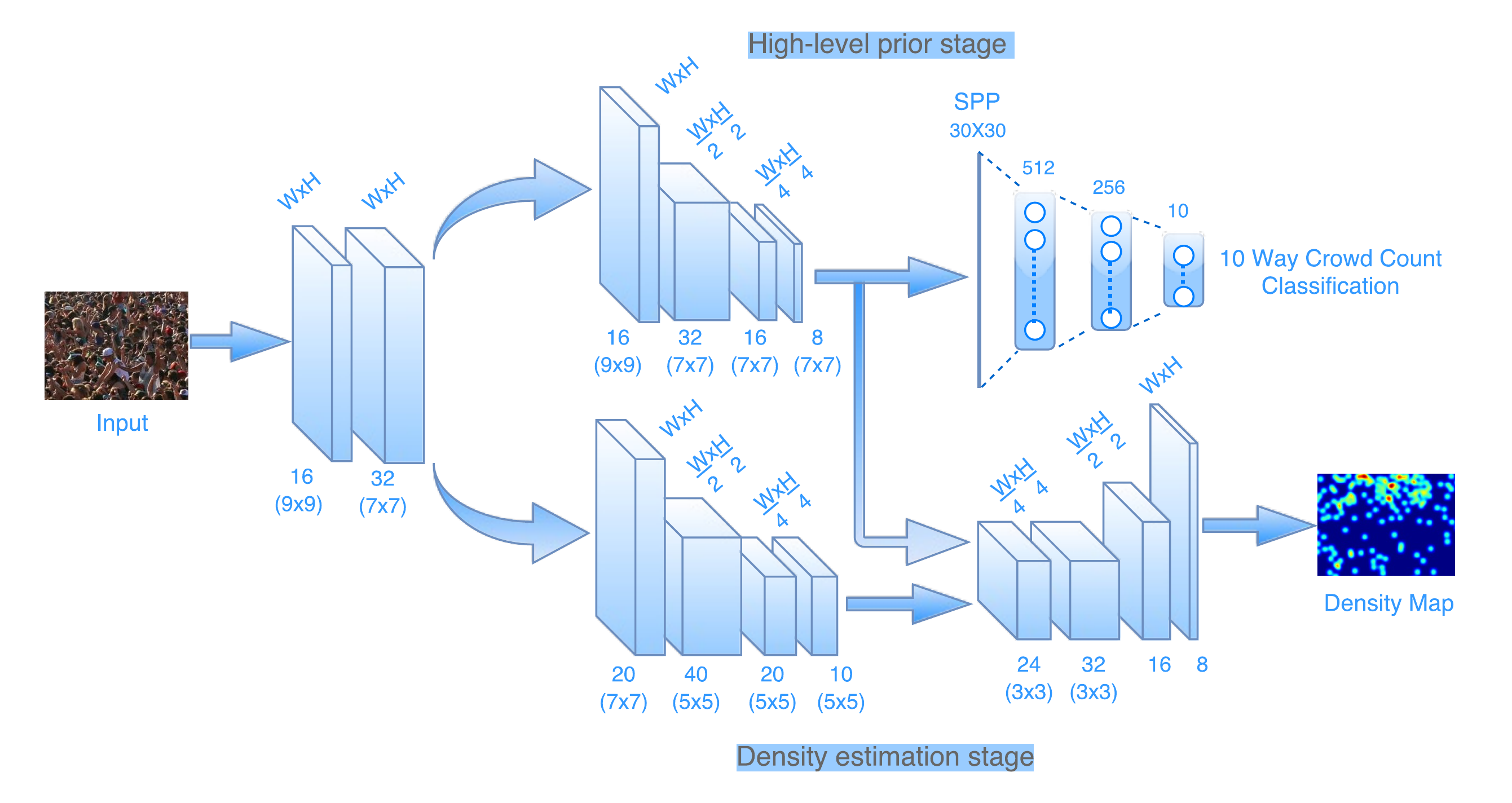}
		\end{center}
		\vskip -16pt \caption{Overview of the proposed cascaded architecture for jointly learning high-level prior and density estimation.}
		\label{fig:arch}
	\end{figure}
	
	\section{Proposed method}
	Inspired by the success of cascaded convolutional networks for related multiple tasks \cite{chen2016cascaded,dai2016instance,ranjan2016hyperface}, we propose to learn two related sub-tasks: crowd count classification (which we call as high-level prior) and density map estimation in a cascaded fashion as shown in Fig. \ref{fig:arch}. The network takes an image of arbitrary size, and outputs crowd density map. The cascaded network has two stages corresponding to the two sub-tasks, with the first stage learning high-level prior and the second stage preforming density map estimation. Both stages share a set of convolutional features. The first stage consists of a set of convolutional layers and spatial pyramid pooling to handle arbitrarily sized images followed by a set of fully connected layers. The second stage consists of a set of convolutional layers followed by fractionally-strided convolutional layers for upsampling the previous layer's output to account for the loss of details due to earlier pooling layers. Two different set of loss layers are used at the end of the two stages, however, the loss of the second layer is dependent on the output of the earlier stage. The following sub-sections discuss the details of all the components of the proposed network.
	
	\subsection{Shared convolutional layers}
	The initial shared network consists of 2 convolutional layers with a Parametric Rectified Linear Unit (PReLU) activation function after every layer. The first convolutional layer has 16 feature maps with a filter size of $9\times9$ and the second convolutional layer has 32 feature maps with a filter size of $7\times7$. The feature maps generated by this shallow network are shared by the two stages: high-level prior stage and density estimation stage. 
	
	\subsection{High-level prior stage}
	Classifying the crowd into several groups is an easier problem as compared to directly performing classification or regression for the whole count range which requires a larger amount of training data. Hence, we quantize the crowd count into ten groups and learn a crowd count group classifier which also performs the task of incorporating high-level prior into the network. The high-level prior stage takes feature maps from the previous shared convolutional layers. This stage consists of 4 convolutional layers with a PReLU activation function after every layer. The first two layers are followed by max pooling layers with a stride of 2.  At the end, the high-level prior stage consists of three fully connected (FC) layers with a PReLU activation function after every layer. The first FC layer consists of 512 neurons whereas the second FC layer consists of 256 neurons. The final layer consists of a set of 10 neurons followed by a sigmoid layer, indicating the count class of the input image. To enable the use of arbitrarily sized images for training, Spatial Pyramid Pooling (SPP) \cite{he2014spatial} is employed as it eliminates the fixed size constraint of deep networks which contain fully connected layers. The SPP layer is inserted after the last convolutional layer. The SPP layer aggregates features from the convolutional layers to produce fixed size outputs and can be fed to the fully connected layers. Cross-entropy error is used as the loss layer for this stage. 
	
	\subsection{Density estimation}
	The feature maps obtained from the shared layers are processed by an another CNN network that consists of 4 convolutional layers with a PReLU activation function after every layer. The first two layers are followed by max pooling layers with a stride of 2, due to which the output of CNN layers is downsampled by a factor of 4. The first convolutional layer has 20 feature maps with a filter size of $7\times7$, the second convolutional layer has 40 feature maps with a filter size of $5\times5$,  the third layer has 20 feature maps with a filter size of $5\times5$ and the fourth layer has 10 feature maps with a filter size of $5\times5$. The output of this network is combined with that of the last convolutional layer of high-level prior stage using a set of 2 convolutional and 2 fractionally strided convolutional layers. The first two convolutional layers have a filter size of $3\times3$ with 24 and 32 feature maps,  respectively. These layers are followed by 2 sets of fractionally strided convolutional layers with 16 and 18 feature maps, respectively. In addition to integrating high-level prior from an earlier stage, the fractionally strided convolutions learn to upsample the feature maps to the original input size thereby restoring the details lost due to earlier max-pooling layers. The use of these layers results in upsampling of the CNN output by a factor of 4, thus enabling us to regress on full resolution density maps. Standard pixel-wise Euclidean loss is used as the loss layer for this stage. Note that this loss depends on intermediate output of the earlier cascade, thereby enforcing a causal relationship between count classification and density estimation.

	\subsection{Objective function}
	The cross-entropy loss function for the high-level prior stage is defined as follows:
	
	\begin{equation}
	\label{eq:lossclass}
	L_c = -\frac{1}{N}\sum_{i=1}^{N}\sum_{j=1}^{M}[(y^i=j)F_c(X_i,\Theta)],
	\end{equation}
	where $N$ is number of training samples, $\Theta$ is a set of network parameters, $X_i$ is the $i^{th}$ training sample, $F_c(X_i,\Theta)$ is the classification output, $y^i$ is the ground truth class and $M$ is the total number of classes.
	
	\noindent The loss function for the density estimation stage is defined as:
	\begin{equation}
	\label{eq:lossdensity}
	L_d = \frac{1}{N}\sum_{i=1}^{N}\|F_d(X_i,C_i,\Theta) - D_{i}\|_2,
	\end{equation}
	where $F_d(X_i, C_i, \Theta)$ is the estimated density map, $D_i$ is the ground truth density map, and $C_i$ are the feature maps obtained from the last convolutional layer of the high-level prior stage.
	
	\noindent The entire cascaded network is trained using the following unified loss function:
	\begin{equation}
	\label{eq:finalloss}
	L = \lambda L_c + L_d,
	\end{equation}
	where $\lambda$ is a weighting factor. 
	
	\noindent This loss function is unlike traditional multi-task learning, because the loss term of the last stage depends on the output of the earlier one.
	
	\subsection{Training and implementation details}
	In this section, details of the training procedure are discussed. To create the training dataset, patches of size $1/4^{th}$ the size of original image are cropped from 100 random locations. Other augmentation techniques like horizontal flipping and noise addition are used to create another 200 patches. The random cropping and augmentation resulted in a total of 300 patches per image in the training dataset. Note that the cropping is used only as a data augmentation technique and the resulting patches are of arbitrary sizes.
	
	Several sophisticated methods are proposed in the literature for calculating the ground truth density map \cite{zhang2015cross,zhang2016single}. We use a simple method in order to ensure that the improvements achieved are due to the proposed method and are not dependent on the sophisticated methods for calculating the ground truth density maps. Ground truth density map $D_i$ corresponding to the $i^{th}$ training patch is calculated by summing a 2D Gaussian kernel centered at every person's location $x_g$ as defined below:
	\begin{equation}
	\label{eq:densitymap}
	D_i(x) = \sum_{{x_g \in S}}\mathcal{N}(x-x_g,\sigma),
	\end{equation}
	where $\sigma$ is the scale parameter of the 2D Gaussian kernel and $S$ is the set of all points at which people are located.
	
	The training and evaluation was performed on NVIDIA GTX TITAN-X GPU using Torch framework \cite{collobert2011torch7}. $\lambda$ was set to 0.0001 in  \eqref{eq:finalloss}. Adam optimization with a learning rate of 0.00001 and momentum of 0.9 was used to train the model. Additionally, for the classification (high-level prior) stage, to account for the imbalanced datasets, the losses for each class were weighted based on the number of samples available for that particular class. The training took approximately 6 hours. 
	
	\section{Experimental results}
	\label{sec:results}
	In this section, we present the experimental details and evaluation results on two publicly available  datasets: ShanghaiTech \cite{zhang2016single} and  UCF\textunderscore CROWD\textunderscore 50 \cite{idrees2013multi}. For the purpose of evaluation, the standard metrics used by many existing methods for crowd counting were used.  These metrics are defined as follows:
	$$
	MAE = \frac{1}{N}\sum_{i=1}^{N}|y_i-y'_i|,\;\;
	MSE = \sqrt{\frac{1}{N}\sum_{i=1}^{N}|y_i-y'_i|^2},
	$$
	where MAE is mean absolute error, MSE is mean squared error, $N$ is number of test samples, $y_i$ is ground truth count and $y'_i$ is estimated count corresponding to the $i^{th}$ sample.

	\subsection{ShanghaiTech dataset}
	
	The ShanghaiTech dataset was introduced by Zhang \etal \cite{zhang2016single} and it contains 1198 annotated images with a total of 330,165 people. This dataset consists of two parts: Part A with 482 images and Part B with 716 images. Both parts are further divided into training and test datasets with training set of Part A containing 300 images and that of Part B containing 400 images. Rest of the images are used as test set. The results of the proposed method are compared with two recent approaches: Zhang \etal \cite{zhang2015cross} and MCNN  by Zhang \etal \cite{zhang2016single} (Table \ref{tab:resultsshanghaitech}). The authors in \cite{zhang2015cross} proposed a switchable learning function where they learned their network by alternatively training on two objective functions: crowd count and density estimation. In the other approach by Zhang \etal in \cite{zhang2016single}, the authors proposed a multi-column convolutional network (MCNN) to address scale issues and a sophisticated ground truth density map generation technique. It can be observed from Table \ref{tab:resultsshanghaitech}, that the proposed method is able to achieve significant improvements without the use of multi-column networks or sophisticated ground truth map generation. Furthermore, to demonstrate the improvements obtained by incorporating high-level prior via cascaded architecture, we evaluated our network without the high-level prior stage (Single stage CNN) on ShanghaiTech dataset. It can be observed from Table \ref{tab:resultsshanghaitech}, that the cascaded learning of count classification and density estimation reduces the count error by a large margin as compared to the single stage CNN.

\begin{figure}[t]
		\centering
		\begin{minipage}{.32\linewidth}
			\centering
			\includegraphics[width=1\linewidth]{figures/sa/IMG_2.jpg}
			\captionsetup{labelformat=empty}
			\captionsetup{justification=centering}
		\end{minipage}
		\begin{minipage}{.32\linewidth}
			\centering
			\includegraphics[width=1\linewidth]{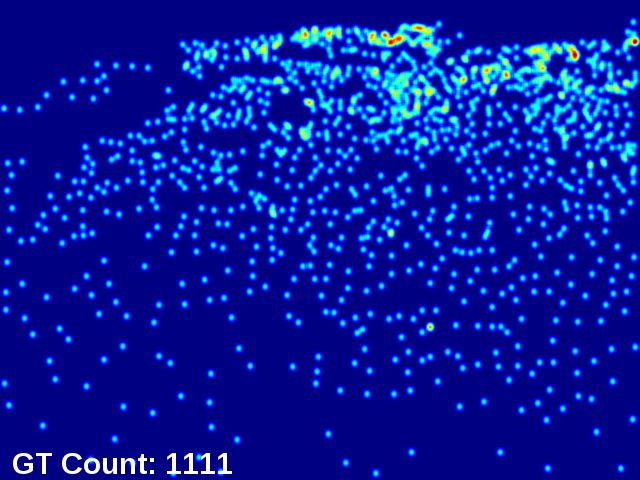}
			\captionsetup{labelformat=empty}
			\captionsetup{justification=centering}
		\end{minipage}	
		\begin{minipage}{.32\linewidth}
			\centering
			\includegraphics[width=1\linewidth]{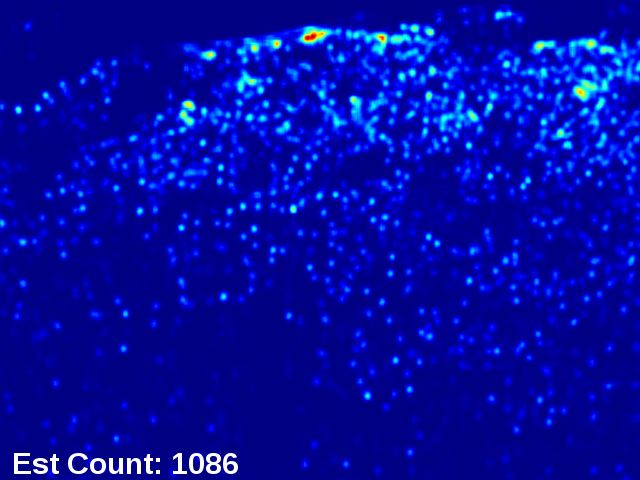}
			\captionsetup{labelformat=empty}
			\captionsetup{justification=centering}
		\end{minipage}
		
		\begin{minipage}{.32\linewidth}
			\centering
			\includegraphics[width=1\linewidth]{figures/sa/IMG_30.jpg}
			\captionsetup{labelformat=empty}
			\captionsetup{justification=centering}
		\end{minipage}
		\begin{minipage}{.32\linewidth}
			\centering
			\includegraphics[width=1\linewidth]{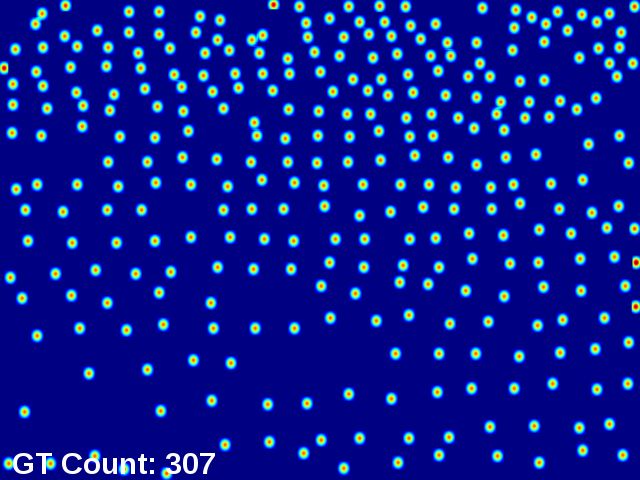}
			\captionsetup{justification=centering}
		\end{minipage}	
		\begin{minipage}{.32\linewidth}
			\centering
			\includegraphics[width=1\linewidth]{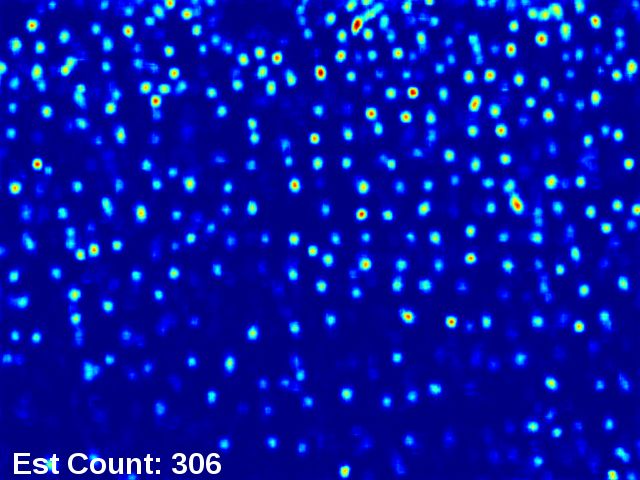}
			\captionsetup{labelformat=empty}
			\captionsetup{justification=centering}
		\end{minipage}
		
		\begin{minipage}{.32\linewidth}
			\centering
			\includegraphics[width=1\linewidth]{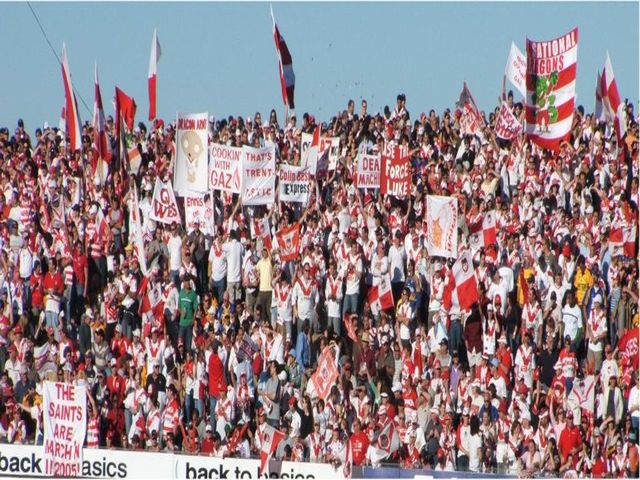}
			\captionsetup{labelformat=empty}
			\captionsetup{justification=centering}
			\caption*{(a)}
		\end{minipage}
		\begin{minipage}{.32\linewidth}
			\centering
			\includegraphics[width=1\linewidth]{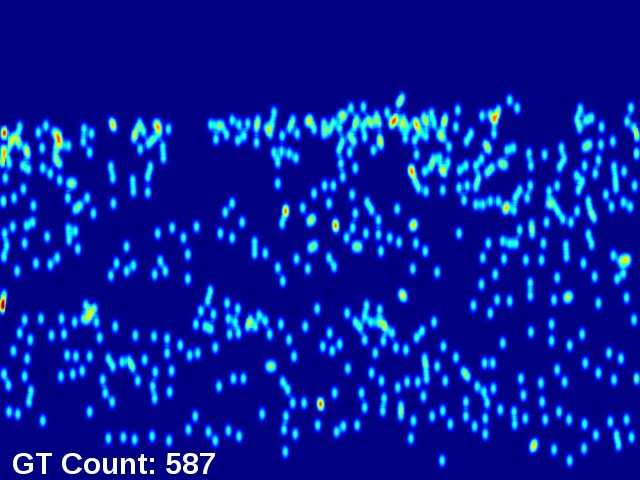}
			\captionsetup{labelformat=empty}
			\captionsetup{justification=centering}
			\caption*{(b)}
		\end{minipage}	
		\begin{minipage}{.32\linewidth}
			\centering
			\includegraphics[width=1\linewidth]{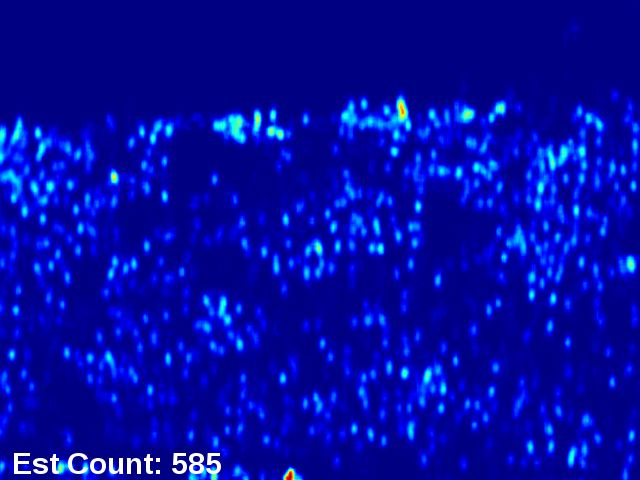}
			\captionsetup{labelformat=empty}
			\captionsetup{justification=centering}
			\caption*{(c)}
		\end{minipage}
		
		\vskip -8pt\caption{Density estimation results using proposed method on ShanghaiTech dataset. (a) Input (b) Ground truth (c) Output.} \label{fig:resultsshanghaitech}
	\end{figure}

	Fig. \ref{fig:resultsshanghaitech} illustrates the density map results obtained using the proposed method as compared to Zhang \etal \cite{zhang2016single} and single stage CNN. It can be observed that in addition to achieving lower count error, the proposed method results in higher quality density maps due to the use of fractionally strided convolutional layers.

	\begin{table}[htp!]
		\caption{Comparison results: Estimation errors on the ShanghaiTech dataset. The proposed method achieves lower error compared to existing approaches involving multi column CNNs and sophisticated density maps.}
		\centering
		\begin{tabular}{|l|c|c|c|c|}
			\hline
			& \multicolumn{2}{c|}{Part A} & \multicolumn{2}{c|}{Part B} \\ \hline
			Method          & MAE          & MSE          & MAE          & MSE          \\ \hline\hline
			Zhang \etal \cite{zhang2015cross}    & 181.8        & 277.7        & 32.0         & 49.8         \\ \hline
			MCNN \cite{zhang2016single}           & 110.2        & 173.2        & 26.4         & 41.3         \\ \hline
			Single stage CNN & 130.4        & 190.9        & 29.3     & 40.5         \\ \hline
			
			Proposed method & \textbf{101.3}        & \textbf{152.4}        & \textbf{20.0}         & \textbf{31.1}         \\ \hline
		\end{tabular}
		\label{tab:resultsshanghaitech}
	\end{table}
	
	\subsection{UCF\textunderscore CC\textunderscore 50 dataset} 
	
\begin{figure}[t!]
		\centering
		\begin{minipage}{.32\linewidth}
			\centering
			\includegraphics[width=1\linewidth]{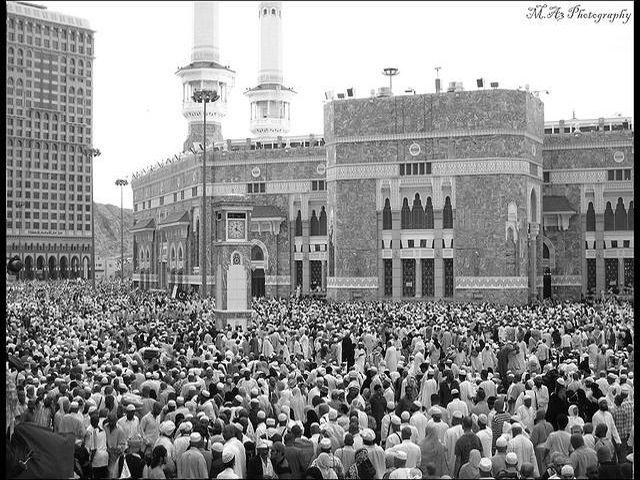}
			\captionsetup{labelformat=empty}
			\captionsetup{justification=centering}
		\end{minipage}
		\begin{minipage}{.32\linewidth}
			\centering
			\includegraphics[width=1\linewidth]{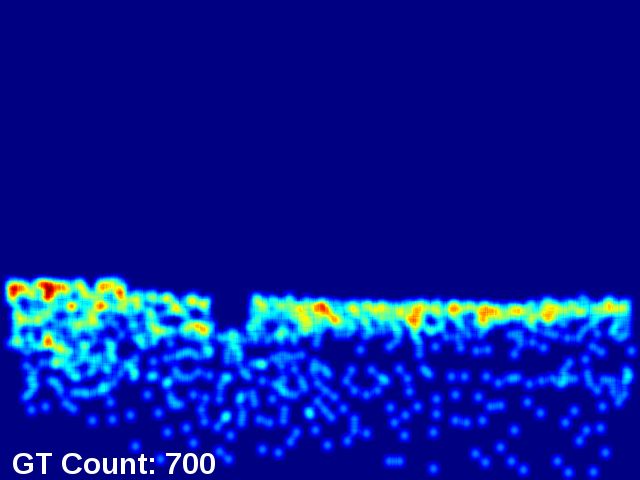}
			\captionsetup{labelformat=empty}
			\captionsetup{justification=centering}
		\end{minipage}	
		\begin{minipage}{.32\linewidth}
			\centering
			\includegraphics[width=1\linewidth]{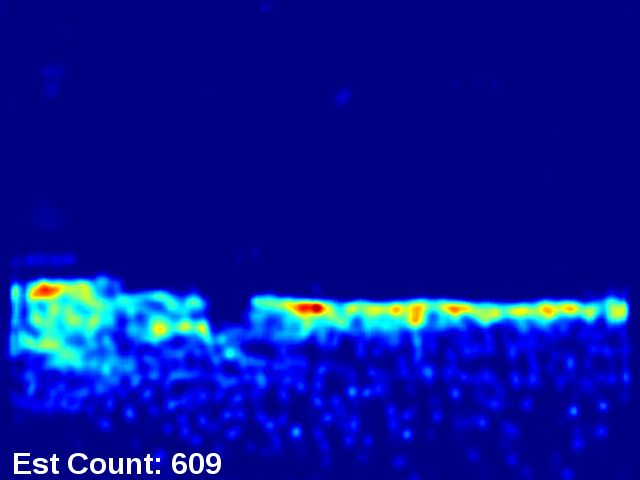}
			\captionsetup{labelformat=empty}
			\captionsetup{justification=centering}
		\end{minipage}
		
		\begin{minipage}{.32\linewidth}
			\centering
			\includegraphics[width=1\linewidth]{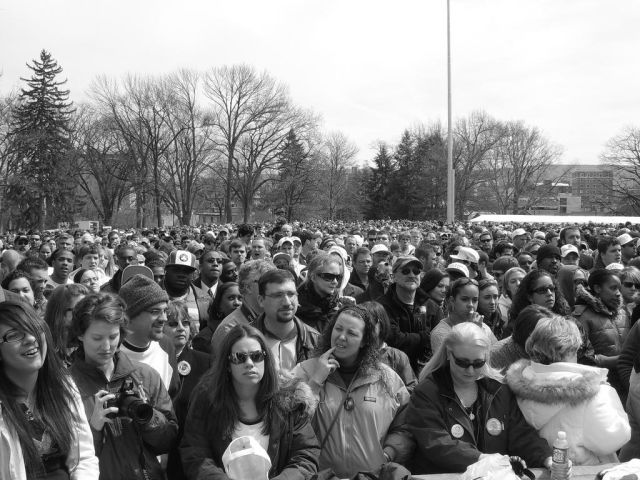}
			\captionsetup{labelformat=empty}
			\captionsetup{justification=centering}
		\end{minipage}
		\begin{minipage}{.32\linewidth}
			\centering
			\includegraphics[width=1\linewidth]{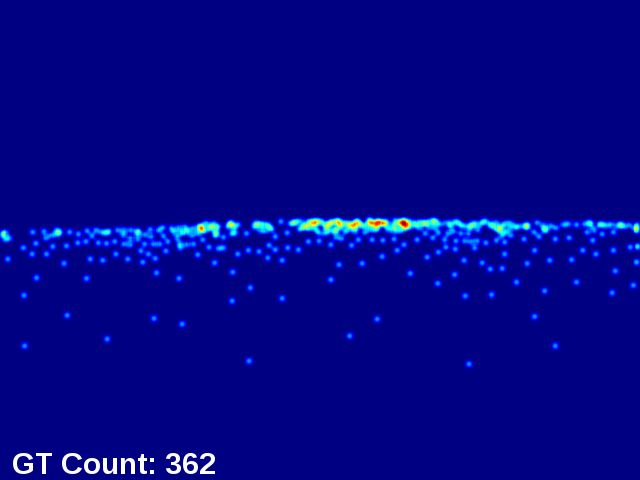}
			\captionsetup{justification=centering}
		\end{minipage}	
		\begin{minipage}{.32\linewidth}
			\centering
			\includegraphics[width=1\linewidth]{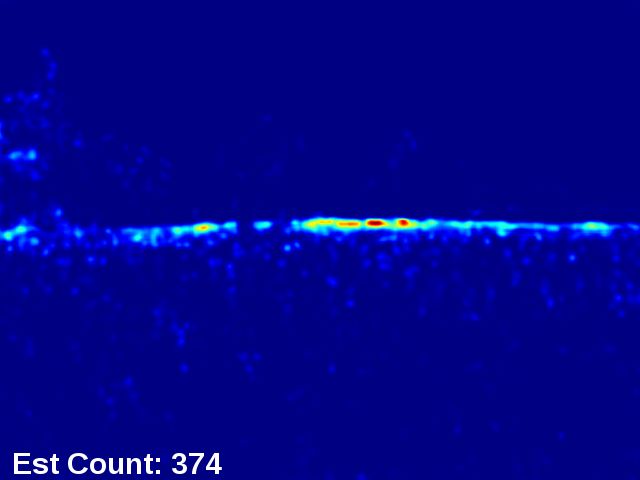}
			\captionsetup{labelformat=empty}
			\captionsetup{justification=centering}
		\end{minipage}
		
		\begin{minipage}{.32\linewidth}
			\centering
			\includegraphics[width=1\linewidth]{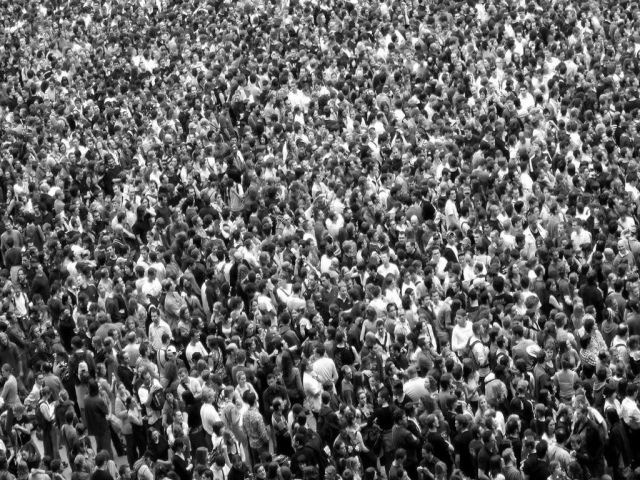}
			\captionsetup{labelformat=empty}
			\captionsetup{justification=centering}
			\caption*{(a)}
		\end{minipage}
		\begin{minipage}{.32\linewidth}
			\centering
			\includegraphics[width=1\linewidth]{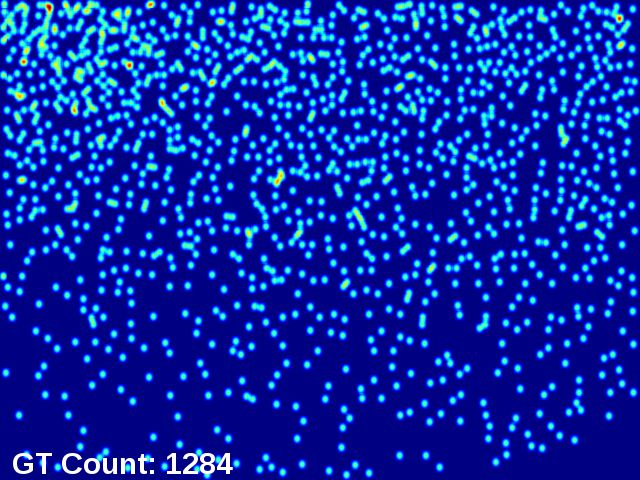}
			\captionsetup{labelformat=empty}
			\captionsetup{justification=centering}
			\caption*{(b)}
		\end{minipage}	
		\begin{minipage}{.32\linewidth}
			\centering
			\includegraphics[width=1\linewidth]{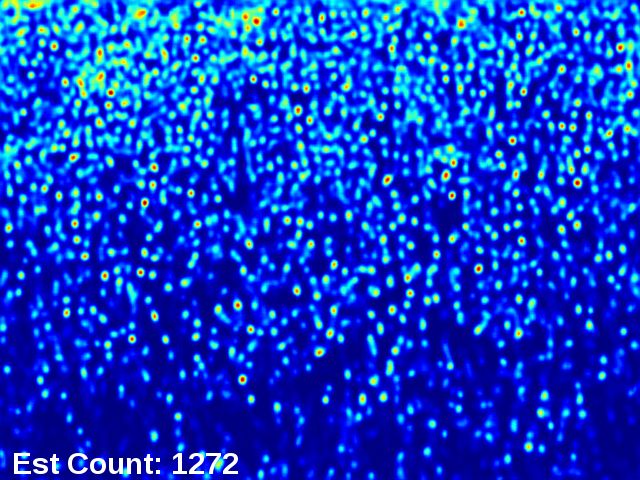}
			\captionsetup{labelformat=empty}
			\captionsetup{justification=centering}
			\caption*{(c)}
		\end{minipage}
		
		\vskip -8pt\caption{Density estimation results using proposed method on UCF\textunderscore CC\textunderscore 50 dataset. (a) Input (b) Ground truth (c) Output.} \label{fig:resultsucf}
	\end{figure}

	The UCF\textunderscore CC\textunderscore 50 is an extremely challenging dataset introduced by Idrees \etal \cite{idrees2013multi}. The dataset contains 50 annotated images of different resolutions and aspect ratios crawled from the internet. There is a large variation in densities across images. Following the standard protocol discussed in \cite{idrees2013multi}, a 5-fold cross-validation was performed for evaluating the proposed method. The results are compared with five recent approaches: Idrees \etal \cite{idrees2013multi}, Zhang \etal \cite{zhang2015cross}, MCNN \cite{zhang2016single}, Onoro \etal \cite{onoro2016towards} and Walach \etal \cite{walach2016learning}. The authors in \cite{idrees2013multi} proposed to combine information from multiple sources such as head detections, Fourier analysis and texture features (SIFT). Onoro \etal in \cite{onoro2016towards} proposed a scale aware CNN to learn a multi-scale non-linear regression model using a pyramid of image patches extracted at multiple scales. Walach \etal \cite{walach2016learning} proposed a layered approach of learning CNNs for crowd counting by iteratively adding CNNs where every new CNN is trained on residual error of the previous layer. It can be observed from Table \ref{tab:resultsucf} that our network achieves the lowest MAE and comparable MSE score. Density maps obtained using the proposed method on sample images from UCF\textunderscore CC\textunderscore 50 dataset are shown in Fig. \ref{fig:resultsucf}.

	\begin{table}[htp!]
		\caption{Comparison results: Estimation errors on the UCF\textunderscore CC\textunderscore 50 dataset.}
		\begin{center}
			\vskip-10pt\begin{tabular}{|l|c|c|}
				\hline
				Method & MAE & MSE \\
				\hline\hline
				Idrees \etal \cite{idrees2013multi}& 419.5 & 541.6 \\
				\hline
				Zhang \etal \cite{zhang2015cross}& 467.0 & 498.5 \\
				\hline
				MCNN \cite{zhang2016single} & 377.6 & 509.1\\
				\hline
				Onoro \etal \cite{onoro2016towards} & 465.7 & 371.8 \\
				\hline
				Walach \etal \cite{walach2016learning} & 364.4  & \textbf{341.4} \\
				\hline
				Proposed method & \textbf{322.8} & 397.9 \\
				\hline
			\end{tabular}
		\end{center}
		
		\label{tab:resultsucf}
	\end{table}
	
	\section{Conclusions}
	
	In this paper, we presented a multi-task cascaded CNN network for jointly learning crowd count classification and density map estimation. By learning to classify the crowd count into various groups, we are able to incorporate a high-level prior into the network which enables it to learn globally relevant discriminative features thereby accounting for large count variations in the dataset. Additionally, we employed fractionally strided convolutional layers at the end so as to account for the loss of details due to max-pooling layers in the earlier stages there by allowing us to regress on full resolution density maps. The entire cascade was trained in an end-to-end fashion. Extensive experiments performed on challenging datasets and comparison with recent state-of-the-art approaches demonstrated the significant improvements achieved by the proposed method.
	
	\section*{Acknowledgement}
	This work was supported by US Office of Naval Research (ONR) Grant YIP N00014-16-1-3134.

	{\small
		\bibliographystyle{ieee}
		\bibliography{egbib}
	}
	
\end{document}